\documentclass[runningheads]{llncs}
\usepackage{graphicx} 
\usepackage{wrapfig}
\usepackage{multirow}
\usepackage{colortbl}

\begin{document}

\title{News about Global North considered Truthful! \\ The Geo-political Veracity Gradient in Global South News}
\author{Sujit Mandava, Deepak P, Sahely Bhadra}
\date{}
\institute{}

\maketitle

\begin{abstract}
While there has been much research into developing AI techniques for fake news detection aided by various benchmark datasets, it has often been pointed out that fake news in different geo-political regions traces different contours. In this work we uncover, through analytical arguments and empirical evidence, the existence of an important characteristic in news originating from the Global South viz., the geo-political veracity gradient. In particular, we show that Global South news about topics from Global North - such as news from an Indian news agency on US elections - tend to be less likely to be fake. Observing through the prism of the political economy of fake news creation, we posit that this pattern could be due to the relative lack of monetarily aligned incentives in producing fake news about a different region than the regional remit of the audience. We provide empirical evidence for this from benchmark datasets. We also empirically analyze the consequences of this effect in applying AI-based fake news detection models for fake news AI trained on one region within another regional context. We locate our work within emerging critical scholarship on geo-political biases within AI in general, particularly with AI usage in fake news identification; we hope our insight into the geo-political veracity gradient could help steer fake news AI scholarship towards positively impacting Global South societies. 
\end{abstract}

\section{Introduction}

It is well understood that popular data sources do not have an even distribution of data from across geo-political regions. In particular, there is often a dominance of Global North contexts within data, often rightly interpreted as a {\it digital exclusion} of other contexts~\cite{graham2022data}. Such data skews and biases have been shown to have significant impacts on computational tasks such as facial recognition~\cite{jaiswal2024breaking} and skin cancer detection~\cite{pope2024skin}. The contours of data biases are complex and different across various sectors and require nuanced analyses to uncover. For example, it has been pointed out that Global South representation in Wikipedia is often due to Global North authors~\cite{graham2022data}; this points to the need to study potential second-order biases that are not visible superficially. 

Data biases are of significant consequence when it comes to media. As noted in~\cite{rambaldi2022review}, when viewing {\it '... media information as a field of knowledge control'}, the development norms from the Global North embed themselves into the public's perception of reality. Fuchs~\cite{fuchs2010new} identifies concentration and transnationalization as tendencies within the media that orient it towards developing a form of information and media imperialism. In a way, the concentration and transnationalization brought about by big news corporations (e.g., Reuters, BBC) covering news from across global regions could dominate the media ecosystems within Global South regions, and shrink the space for media that could provide organic narratives. The rise of social media, virtually all driven out of Global North (or more specifically, Silicon Valley), has added to the pace of Global North dominance. \cite{chinmayi2020ai} documents one such instance, where the friction between the design choices of Facebook and the social realities of Myanmar resulted in severe consequences during the Rohingya crisis. There are some notable challenges to such Global North hegemony such as Al-Jazeera~\cite{seib2005hegemonic}, but these remain few and far between.  

An aspect relating to media, the phenomenon of fake news, has been consistently ranked as among the top global risks\footnote{https://www.weforum.org/press/2024/01/global-risks-report-2024-press-release/} within contemporary society. This was particularly highlighted during the course of the 2016 US election~\cite{grinberg2019fake}, and arguably has only intensified since. There has been tremendous interest in the AI community in responding to this challenge, and fake news detection has emerged as a burgeoning area of AI research~\cite{iqbal2023relationship}. Yet, due to issues such as data and media bias highlighted above, the AI capacity for fake news detection is aligned better with the needs of the Global North than those of the Global South. The complex contours of such biases have attracted recent attention, with one study highlighting the geo-political bias relating to the usage of emotions within techniques to combat misinformation~\cite{deepak2024geo}. It is imperative to uncover the points of injection of such geo-political biases to fuel efforts that seek an equitable impact of AI methods in resisting disinformation across global societies. We locate our work against the backdrop of this larger goal. 

The focus of this paper is the characterization of a consistent trend in Global South news, one that we call the {\it geo-political veracity gradient}. We start by outlining this pattern formally as a conjecture, followed by three types of analyses of this trend. First, we outline factors from the political economy of media that could explain this trend. Second, we quantitatively outline the extent of this trend through empirical investigations of public datasets. Third, we empirically uncover the consequences of this trend for cross-regional usage of AI-based fake news detection. We conclude by drawing attention to the need to factor this trend into building AI-based methods for fake news detection. 


\section{Geo-political Veracity Gradient in Global South News}\label{sec:conjecture}

We outline the key insight of this paper as a conjecture. 

\vspace{0.1in}
{\it When considering news originating from the Global South, alignment with topics that pertain to the Global North correlates with higher degrees of veracity. }
\vspace{0.1in}

As an example, the conjecture suggests that if an India-based news source\footnote{we use the term news source in a generic manner, so it includes posts from reputed media as well as popular social media users or influencers} reports on events in the US, it is more likely to be truthful reporting. In other words, if we split news from Global South into two distinct subsets viz., those on topics relating to Global North, and those solely reporting on Global South, the conjecture suggests that it is likely that the latter would have a higher prevalence of misinformation than the former. While we present this as a generalized conjecture, we do recognize that the categories of Global North and Global South encompass huge amounts of diversity, and the pattern stated above may not hold for certain combinations. 

\section{Political Economy of Fake News}

We now provide argumentative support for the conjecture based on insights from the political economy of media and fake news. 

We begin with some background on contemporary media. Ad-driven media such as online and social media, which are increasingly becoming mainstream avenues for news consumption, are primarily governed by the notion of the {\it attention economy}. This involves the characterization of {\it 'human attention as a scarce but quantifiable commodity'}~\cite{crogan2012paying}; this attention commodity is eventually monetized through advertising models, resulting in a proliferation of ads in online life. With the commodity form of attention being key, the forces of news creation are swayed to work in ways such that their news can amass as much attention commodity as possible. This attention incentive progressively crowds out other factors such as news quality, leading to an increasing prevalence of sensationalized~\cite{hendriks2018proving} and clickbait-style content within our news streams. 

\begin{figure}
\vspace{-0.3in}
  \includegraphics[width=\linewidth]{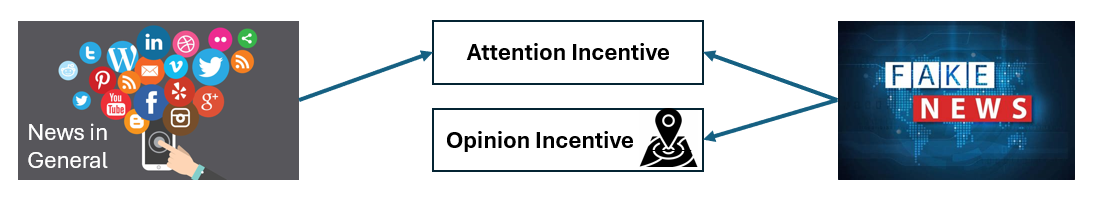}
  \caption{Incentives in General News vis-a-vis Fake News: A Simplified View}
  \label{fig:structure}
  \vspace{-0.2in}
\end{figure}

The category of fake news enhances monetization by opening up an additional pathway viz., {\it direct monetization of content creation}. For example, a content creator could be paid by a movie producer for promoting their film, or by a competitor to slander the same. News writers could be funded to flatter or defame products or political actors. The financial motive stems from the urge to use fake news to influence opinions among people in the real world. We call this the {\it opinion incentive} in fake news. 

The argument so far is summarized in Fig~\ref{fig:structure}. We now turn our attention to a qualitative distinction between the two incentives. First, we must look at the geo-localization of media as a background. A news source would naturally tend to attract audiences associated with its region or related regions. This is because news (or social media posts) is often about reporting on events in the physical world, and a local presence allows deeper local coverage, thus attracting more local audiences and enabling a positive feedback loop. Such localization of media {\it audiences} does not necessarily predicate a localization of general news {\it content}. There may be sufficient local interest (i.e., attention potential) in a geographically remote event. For example, when there is interest locally, Chinese media may post on the US elections, and Latin American movie buffs may report on Hollywood movies. However, when it comes to the opinion incentive, such remote coverage becomes less justified. For example, it would serve little purpose to fund fake news about a US presidential candidate within Chinese media, and any defamation of a Hollywood movie within Latin America would cause only a blip of damage, if at all any. Thus, fake news, due to being influenced by the opinion incentive, tends to be much more anchored to the local remit of the source, an effect indicated by the pin icon in Fig~\ref{fig:structure}. {\it This makes fake news much more geo-localized to the regional remit of the source than news in general}.

Considering Global South and Global North as regions, the above reasoning could yield two kinds of outcomes viz. the conjecture in Section~\ref{sec:conjecture}, and vice versa. The latter suggests that news about the Global South from Global North sources may be of higher veracity. However, the economic, political, and cultural hegemony of the Global North~\cite{globalnorth}, underpinned by the long colonial era, reflects in contemporary media as an interest towards Global North topics within the Global South, one that is not matched the other way to the same extent. Thus, the mirror image of the conjecture, while theoretically plausible, is not likely a significant or widespread effect. 

\section{Empirical Evidence for Geo-political Veracity Gradient}

\begin{wraptable}{r}{8.5cm}
\vspace{-0.6in}
\begin{center}
\begin{tabular}{ |c|p{1.2cm}|p{1.2cm}|p{1.2cm}|p{1.2cm}|} 
  \hline
  {\bf Word/Phrase} & {\bf GN-R} & {\bf GN-F} & {\bf GS-R} & {\bf GS-F} \\
  \hline
  \hline
  United States & 28.76\% & 17.72\% & 5.08\% & 1.23\% \\
  Trump & 44.83\% & 53.34\% & 4.55\% & 2.44\% \\
  Clinton & 10.17\% & 21.92\% & 0.48\% & 0.15\% \\
  Britain & 7.16\% & 1.04\% & 1.34\% & 0.27\% \\
  Washington & 18.19\% & 13.99\% & 3.2\% & 0.75\% \\
  Europe & 5.92\% & 2.79\% & 2.56\% & 0.48\% \\
  Japan & 3.66\% & 0.52\% & 2.5\% & 0.81\% \\
  \hline
\end{tabular}
\caption{Frequency of Global North Words/Phrases}
\label{tab:freq}
\end{center}
\vspace{-0.4in}
\end{wraptable}

Towards providing empirical evidence for the conjecture in Section~\ref{sec:conjecture}, we assemble a dataset with four parts; each part containing real or fake news from either Global South or Global North. These are indicated as GN-R, GN-F, GS-R and GS-F. While the Global North dataset is simply the popular ISOT dataset\footnote{{\tiny https://onlineacademiccommunity.uvic.ca/isot/2022/11/27/fake-news-detection-datasets/}}, the Global South dataset is formed by news datasets in Indian contexts found on Kaggle\footnote{https://www.kaggle.com/datasets/kuberiitb/indian-news-articles}\footnote{https://www.kaggle.com/datasets/banuprakashv/news-articles-classification-dataset-for-nlp-and-ml}, along with the FakeNewsIndia dataset\cite{DHAWAN2022130}. These have been randomly downsampled to ensure that each of the four partitions has similar counts. 

Table~\ref{tab:freq} illustrates the frequency of some few notable entities of interest in Global North spheres across the four data partitions, frequency measured as the fraction of articles containing the word/phrase. While most words are Global North words, their high frequencies within GN-R and GN-F confirm their high prevalence in GN data. The key observation from the table is the sharp deterioration of the frequency of these words as we move from GS-R to GS-F; this supports our conjecture. The words' frequency in GS-R ranges from around two to five times that of their frequencies in GS-F, indicating the intensity of the trend. Our findings are contingent on the frequentist statistics we use; future work could delineate the trend further by leveraging more sophisticated NLP techniques. 

\section{Consequences for AI-based Fake News Detection}

Our conjecture on the geo-political veracity gradient in Global South news could throw light on the frictions experienced while applying fake news detection models outside their regions of production. We outline and empirically illustrate two such effects herein, using FNDNet~\cite{kaliyar2020fndnet} for fake news classification. 

\begin{table}[!htb]
    \vspace{-0.2in}
    \begin{minipage}{.5\linewidth}
     \centering
      \caption{GN-trained AI on GS-test}
        \begin{tabular}{|c|p{1cm}|p{1cm}|}
            \hline
            Actual & \multicolumn{2}{|c|}{Predicted} \\
            \cline{2-3}
            Labels & Fake & Real \\
            \hline
            Fake & \cellcolor{yellow} 455 & \cellcolor{yellow} 482 \\
            \hline
            Real & 285 & 605 \\
            \hline
        \end{tabular}
        \label{tab:gntraingstest}
    \end{minipage}%
    \begin{minipage}{.5\linewidth}
      \centering
        \caption{GS-trained AI on GN-test}
        \begin{tabular}{|c|p{1cm}|p{1cm}|}
            \hline
            Actual & \multicolumn{2}{|c|}{Predicted} \\
            \cline{2-3}
            Labels & Fake & Real \\
            \hline
            Fake & 90 & \cellcolor{yellow}847 \\
            \hline
            Real & 6 & \cellcolor{yellow} 884 \\
            \hline
        \end{tabular}
        \label{tab:gstraingntest}
    \end{minipage} 
    \vspace{-0.2in}
\end{table}

Consider an AI model for fake news classification trained using Global North data, being applied within Global South contexts. With the pervasive availability of pre-trained/foundation models in AI, this is a very likely scenario. The model, due to being trained on Global North data, may have internalized patterns of lexical correlations between Global North words and real/fake labels. These patterns would be of low utility in determining labels for Global South fake news, which has a very sparse presence of Global North words. Thus, such a model is likely to produce a lot of misclassifications of fake news as real (i.e., {\it false negatives}) and is likely to be incompetent for decision-making on Global South fake news data.


Let's now consider the less likely\footnote{practically less likely due to Global North hegemony in computing and AI.} analogous case, that of taking fake news AI trained on Global South data, and applying it within Global North contexts. The AI would likely have identified the utility of Global North words as a way to distinguish between real and fake labels in its Global South training data. In particular, it would have internalized a high propensity to choose the label {\it real} while encountering Global North words. Now, when applied in Global North contexts, this would translate to a high inclination towards {\it real} label in general, with all news in Global North contexts naturally exhibiting a high prevalence of Global North words. 

We observed the above-expected effects consistently in our empirical results. Tables~\ref{tab:gntraingstest} and~\ref{tab:gstraingntest} plot the confusion matrices in the respective cases. The dismal performance of the models over fake news in the former and the high propensity towards predicting as real in the latter are both highlighted in the tables. 

\section{Discussions and Conclusions}

\noindent{\bf Discussion:} Scholarship in data-driven algorithms and AI, in general, has been facilitated by a high prevalence of benchmark datasets. The production of benchmark datasets has been increasingly seen as a service to the community while also offering high citation-gathering incentives. However, this trend promotes a one-size-fits-all ethos, which when read against the backdrop of Global North hegemony makes such benchmark datasets largely Global North-centric. This works out to codify and reinforce a neglect of Global South complexities and nuances in AI scholarship. Any efforts to bring the benefits of AI to the Global South, especially when it comes to highly sensitive and consequential realms such as fake news detection, ought to be accompanied by an urge to understand, contextualize, and assimilate the unique characteristics of Global South contexts within algorithms. Within the realm of fake news, extant scholarship asserts the distinctive role of affect in Global South contexts~\cite{deepak2024geo}, whereas our work uncovers the unique impacts of geo-localized words in Global South. We do expect that there would be many more fine-grained distinctive factors at the national or regional level within the Global South, ones that may require much more investigation to uncover. 

\noindent{\bf Conclusions:} In this paper, for the first time, we analyzed the impact of regionally-oriented words and their influence on fake news detection within the Global South. In particular, we outlined a consistent pattern within Global South news, where the prevalence of Global North words is directly related to news veracity, one we define as the geo-political veracity gradient. We provide three approaches to support this conjecture. First, we provided argumentation grounded on the political economy of media and fake news as analytical support for the conjecture. Second, we analyzed popular datasets and illustrated the empirical measurability and intensity of the pattern. Third, we illustrated the consequences of this veracity gradient in limiting the cross-regional application of AI for fake news detection. By providing these arguments, we seek to bring attention to the nuances of fake news in different regions, particularly with respect to the Global South, to achieve an equitable impact in the larger objective of tackling disinformation with AI methods. 

\noindent{\bf Future Work:} We find that a lot of marginalized communities within the Global South are not represented as much within digital media as compared to traditional media. In future work, we intend to consider how such representational gradients would influence AI for fake news detection. 

\bibliographystyle{splncs04}
\bibliography{refs}

\end{document}